\documentclass[journal]{IEEEtran}
\usepackage{amssymb}
\usepackage{graphicx}
\usepackage{subfigure}
\usepackage{float}
\usepackage{amsmath}
\usepackage{booktabs}
\usepackage[numbers,sort&compress]{natbib}

\newcommand{\grad}{\ensuremath{^{\circ}}}
\begin{document}
%
\title{Mapping Temporal Variables into the NeuCube for Improved Pattern Recognition, Predictive Modelling and Understanding of Stream Data}
%
%
%

\author{Enmei Tu,
        Nikola Kasabov,~\IEEEmembership{Fellow,~IEEE,}
        and Jie Yang
\thanks{Enmei Tu and Jie Yang are with the Institute of Image Processing and Pattern Recognition, Shanghai Jiao Tong University.}
\thanks{Nikola Kasabov is with the Knowledge Engineering and Discovery Research Institute, Auckland University of Technology.}
\thanks{This manuscript is submitted to Special issue on Neurodynamic Systems for Optimization and Applications}}

%
%

\markboth{Journal of IEEE Trans. on NNLS Special issue on Neurodynamic Systems for Optimization and Applications}%
{Tu \MakeLowercase{\textit{et al.}}: Mapping, Learning and Predictive Data Modeling}
%



\maketitle

\begin{abstract}
This paper proposes a new method for an optimized mapping of temporal variables, describing a temporal stream data, into the recently proposed NeuCube spiking neural network architecture. This optimized mapping extends the use of the NeuCube, which was initially designed for spatiotemporal brain data, to work on arbitrary stream data and to achieve a better accuracy of temporal pattern recognition, a better and earlier event prediction and a better understanding of complex temporal stream data through visualization of the NeuCube connectivity. The effect of the new mapping is demonstrated on three bench mark problems. The first one is early prediction of patient sleep stage event from temporal physiological data. The second one is pattern recognition of dynamic temporal patterns of traffic in the Bay Area of California and the last one is the Challenge 2012 contest data set. In all cases the use of the proposed mapping leads to an improved accuracy of pattern recognition and event prediction and a better understanding of the data when compared to traditional machine learning techniques or spiking neural network reservoirs with arbitrary mapping of the variables.
\end{abstract}


%
\IEEEpeerreviewmaketitle

\section{Introduction}
\IEEEPARstart{T}{emporal} data have been collected in various fields, such as brain science, ecology, geophysics, social sciences. Temporal data may contain complex temporal patterns that would need to be learned and extracted in a computational model. In some cases, the variables describing the temporal data have spatial attributes, e.g. the 3D location of the channels in EEG data \cite{kasabov2014neucube} and the patterns that need to be learned become spatiotemporal. Learning dynamic patterns from temporal and spatiotemporal data is challenging task as the temporal features may manifest complex interaction that also may change dynamically over time. The ‘time-windows’ of important temporal or spatiotemporal patterns may change over time which is seldom covered by traditional machine learning methods such as regression techniques, support vector machines (SVM) or multi-layer perceptrons (MLP).


Many recurrent models have been proposed to learn spatiotemporal relationship from signal, such as the recursive self-organizing network models \cite{voegtlin2002recursive, hammer2004recursive}, recurrent neural network \cite{mikolov2011extensions, sutskever2011generating}. Recently NeuCube \cite{kasabov2012neucube, kasabov2014neucube} has been proposed to capture time and space characteristics of spatiotemporal brain data in a spiking neural network (SNN) architecture. The NeuCube architecture consists of: input data encoding module, that encodes multivariable continuous temporal stream data into spike trains; a 3D recurrent SNN cube (SNNcube) where input data is mapped and learned in an unsupervised mode; a SNN classifier that learns in a supervised mode to classify spatiotemporal patterns of the SNNcube activities that represent patterns from the input data. The effectiveness and superiority of NeuCube to model brain data (such as EEG)  has been demonstrated in previous works. \cite{kasabov2015spiking} studies the modelling ability of NeuCube for electroencephalography  (EEG) spatiotemporal data measuring complex cognitive brain processes. In  \cite{taylor2014feasibility} NeuCube is used for modelling and recognition of complex EEG spatiotemporal data related to both physical and intentional (imagined) movements. \cite{doborjeh2014classification, doborjeh2015dynamic} presents the results of using NeuCube to classify/cluster fMRI data and \cite{capecci2015modelling} uses NeuCube for the analysis of EEG data recorded from a person affected by Absence Epileptic. \cite{capecci2015analysis} studies how NeuCube can help understand brain functional changes in a better way.

While in these studies, NeuCube is used to model brain data and the features are manually mapped into the system according to the spatial location where the signals are sampled from, in this paper we present a fully automatic input mapping method based on graph matching technique that enables NeuCube to model any spatiotemporal data and leads to better spatiotemporal pattern recognition, early event prediction and model visualization and understanding.

The main contributions of this paper are as follows:
\begin{itemize}
  \item We propose a new graph matching algorithm to optimize the mapping of the input features (variables) of arbitrary temporal stream data into a 3D SNNcube with the goal to achieve a better pattern recognition accuracy, an earlier event prediction from temporal data and a better understanding of the data through model visualization.
  \item We proposed a graph based  semi-supervised learning algorithm to automatically analyse the neuronal cluster structure inside the trained SNNcube and develop various dynamic functions for visualization  of the neuronal activity states and synaptic evolving progress during learning process and we prove the convergence of the new algorithm.
  \item  We present two algorithms, spike density correlation and  maximum spike coincidence, for spike trains similarity measure that can be used in the graph matching algorithm for two  commonly encountered types of spike trains: bipolar spike trains and unipolar spike trains.
  \item We introduce an Adaptive Threshold Based (ATB)  encoding algorithm by using  mean and standard deviation information of the signal gradient  to calculate its threshold and thus make ATB self-adaptive to signal changes.
  \item We develop an optimization  component using genetic algorithm that can automatically optimize all  NeuCube  parameters in once and thus enable users to find optimal parameters' value easily and achieve good results.
\end{itemize}

The remainder of this paper is organized as follows. Section 2 describes the NeuCube architecture. Section 3 proposes a new method for the optimization of the mapping of the input temporal features into the SNNcube. Experiments on three benchmark data sets are reported in Section 4, followed by conclusions and discussions in Section 5.

\section{The NeuCube Architecture}
\begin{figure}[ht]
  \centering
  \includegraphics[width=6cm]{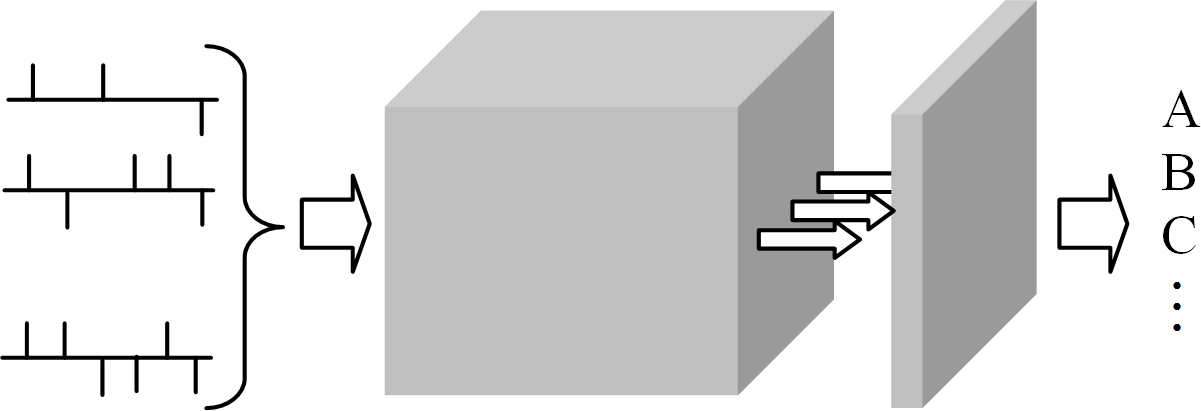}

  \caption{A simplified representation of the NeuCube architecture \cite{kasabov2014neucube, kasabov2012neucube}. Spike trains (left) are fed to the SNNcube and the neuron firing states vectors of the SNNcube are used to train a SNN classifier.}\label{architecture}
\end{figure}

The main parts of NeuCube are: input encoding module; a three-dimensional recurrent SNN reservoir/cube (SNNcube); an evolving SNN classifier. Fig. \ref{architecture} shows the block diagram of the NeuCube architecture. The SNNcube is a scalable module. Its size  is controlled by three parameters: ${{n}_{x}},\ \,{{n}_{y}}$ and${{n}_{z}}$, representing the neuron number along $x$, $y$ and $z$ axes, respectively.  The total number of neurons in the SNNcube is $N={{n}_{x}}\times {{n}_{y}}\times {{n}_{z}}$. As a first implementation here, we use a probabilistic leaky integrate and fire model (LIFM) \cite{kasabov2010spike}.

The encoding module converts continuous data streams into discrete spike trains, suitable to be processed in the SNNcube, because spike neural networks can only process discrete spike trains. After encoding, the information contained in original continuous signal is transformed into forms of spikes.

The NeuCube is trained in a two-stage learning process. The first stage is unsupervised learning that makes the SNNcube learn spatiotemporal relations from the input data by adjusting the connection weights in SNNcube. The second stage is supervised learning that aims at learning the class information associated with each training spatiotemporal sample.

To be more specific, the unsupervised learning stage is intended to encode ‘hidden’ spatiotemporal relationships from the input data into neuronal connection weights. According to the Hebbian learning rule, if the interaction between two neurons is persistent, then the connection between them will be strengthened. In particular, we train the SNNcube using spike-timing dependent synaptic plasticity (STDP) learning rule \cite{song2000competitive}: if neuron $j$ fires before neuron $i$, then the connection weight from neuron $j$ to neuron $i$ will increase and, if the spikes are in a reverse order, the connection from neuron $i$ to neuron $j$ will decrease . This ensures that the time difference in the input spiking trains, which encode the temporal patterns in the original input signals, will be captured by the neuron firing state and the unsymmetrical connection weights in the reservoir.
%

The second training stage is to train an output classifier using class label information associated with the training temporal samples. The dynamic evolving Spike Neural Networks (deSNN) \cite{kasabov2013dynamic} is employed here as an output classifier, because deSNN is computationally efficient and emphasizes the importance of both first spikes arriving at the neuronal inputs (observed in biological systems \cite{thorpe1998rank}) and the following spikes (which in some stream data are more informative).

Once a NeuCube model is trained, all connection weights in the SNNcube and in the output classification layer are established. These connections and weights can change based on further training (adaptation), because the evolvable characteristic of the architecture.

\section{A Method for an Optimal Mapping of Temporal Input Variables into a 3D SNNcube based on the Graph Matching Algorithm}
\subsection{A method for mapping input temporal variables into a 3D SNNcube based on graph matching optimisation algorithm  }
Given a particular spatiotemporal data set, it is important to optimise the mapping of the data into the 3D SNNcube for an optimal learning a better understanding of the spatiotemporal patterns in the data. For some spatiotemporal data, such as brain EEG, there is prior information about the location of each channel (input feature) and this information can be readily utilized for mapping the EEG temporal signal into the SNNcube \cite{kasabov2012neucube}. But for other common applications such as climate temporal data, we do not have such spatial mapping information. And the way temporal data is mapped into the SNNcube would significantly impact the results.  Here we introduce a new method to map temporal  variables into the SNNcube for a better pattern recognition, a better and earlier event prediction and a better visualisation of the model to explain the data.


Suppose there are $s$ temporal samples in the data set, measured through $v$ temporal variables and the observed time length of each sample is $t$. We first choose randomly $v$ input neurons from the SNNcube. Then we map the variables into the SNNcube following the principle:\emph{ input variables/features that, after the input data transformation into spike trains, represent highly correlated spike trains are mapped to nearby input neurons}. Because high correlation indicates that the variables are likely to be more time dependent with each other, and this relationship should also be reflected in the connectivity of the 3D SNNcube. Spatially close neurons in the SNNcube will capture in their connections more temporal interactions between the input variables mapped into these neurons.  The principle of mapping similar input vectors into topologically close neurons is known from the SOM \cite{kohonen1995springer}, but in SOM these are static vectors and the similarity is measured by the Euclidean distance. Now, we address the problem of mapping temporal sequences, rather than static vectors, into a 3D SNN.

Specifically, we construct two weighted graphs: the input neuron similarity graph (NSG) and the time series/signals similarity graph (SSG). In the NSG, the input neurons’ spatial 3D coordinates are denoted by ${{V}_{NSG}}=\left\{ ({{x}_{i}},{{y}_{i}},{{z}_{i}})|i=1..v \right\}$ and the graph edges are determined in the following way: each input neuron is connected to its $k$ nearest input neurons and the edges are weighted by the inverse of the Euclidean distance between them.

In the SSG, we denote the spike train corresponding to the feature $i$ as
 \begin{equation}
 {{s}_{i}}=\sum\limits_{k=1}^{{{N}_{i}}}{\delta (t-{{t}_{k}})}
 \end{equation}
 where $\delta(t)$ is the Dirac delta function. The set of spike trains corresponding to all the features is  ${{V}_{SSG}}=\left\{ {{s}_{i}}|i=1..v \right\}$ and it forms the graph vertex set. The graph edges are constructed in this way: each spike density function is connected to its $k$ highest correlated neighbours and the edges are weighted by the similarity between them. Measuring the similarity between spike trains is an important problem in neural science and have been studied for a long history \cite{van2001novel, dauwels2009similarity}. Here we propose two simple but effective spike trains similarity measuring algorithms for two  most commonly encountered spike train types, respectively. The first one is spike density correlation, which is suitable for measuring unipolar  spike trains with two state (firing or unfiring). This kind of spike train is commonly observed in glutamatergic neurons and produced by some spike encoding  algorithms such as  Bens Spiker Algorithm (BSA) \cite{schrauwen2003bsa}. The other one is maximum spike coincidence, which is suitable for measuring  bipolar spike trains with three state (excitatory, rest or inhibitory). This kind of spike train is commonly produced by serotonergic neurons, as well as some encoding algorithms such as address event representation \cite{dhoble2012line}.

\subsubsection{spike density correlation}
\begin{figure}[ht]
  \centering
  \includegraphics[width=6cm]{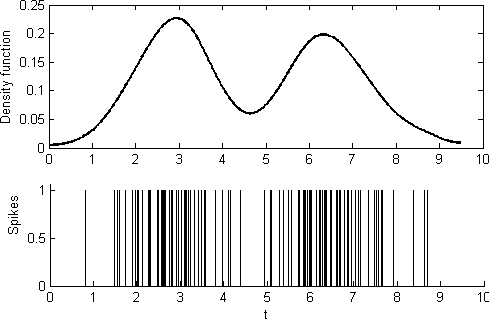}

  \caption{Spike density estimation result.}\label{SDE}
\end{figure}

Given a unipolar spike train $si$, we first use kernel density estimation method to estimate its spike density function  by
 \begin{equation}
{{p}_{i}}(t)=\frac{1}{{{N}_{i}}}{{K}_{h}}(t)*\sum\limits_{k=1}^{{{N}_{i}}}{\delta (t-{{t}_{k}})}=\frac{1}{{{N}_{i}}h}\sum\limits_{k=1}^{{{N}_{i}}}{K(\frac{t-{{t}_{k}}}{h})}
\end{equation}
 where $*$ is the function convolution operator and $K(t)$ is a kernel function, with kernel bandwith $h$. Fig. \ref{SDE} displays the spike density estimation result.
Thereafter, the similarity between spike train $s_i$ and $s_j$ is defined as the Pearson's linear correlation coefficient
 \begin{equation}
 \label{SpikeSim1}
 {{\omega }(s_i, s_j)}=\frac{E\left[ \left( X-{{\mu }_{X}} \right)\left( Y-{{\mu }_{Y}} \right) \right]}{{{\sigma }_{X}}{{\sigma }_{Y}}}
 \end{equation}
 where $\mu_{X}$ and $\sigma_X$ are the mean and the standard deviation of $X$, respectively. $E$ is the expectation. Here $X$ and $Y$ are the random variables corresponding to $s_i$ and $s_j$, respectively.
 \subsubsection{maximum spike coincidence}
\begin{figure}[ht]
  \centering
  \includegraphics[width=6.5cm]{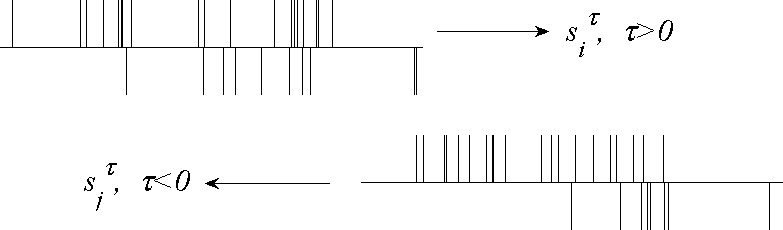}

  \caption{The process of computing maximum spike coincidence.}\label{MSC}
\end{figure}

Given two bipolar spike trains $s_i$ and $s_j$, we define the similarity between them as the maximal number of spikes that coincide while one spike train moves over the other spike train
\begin{equation}
 \label{SpikeSim2}
{{\omega }(s_i, s_j)}=\underset{\tau }{\mathop{\max }}\,\sum\limits_{k=1}^{{{N}_{i}}}{I\left( {{s}_{i}}(k),s_{j}^{\tau }(k) \right)}
\end{equation}
where $s_{i}^{\tau }$ is the result of translating $s_i$ with $\tau$ time unit, i.e.
 \begin{equation}
 {{s}_{i}}=\sum\limits_{k=1}^{{{N}_{i}}}{\delta (t-\tau-{{t}_{k}})}
 \end{equation}
Note that while $\tau$ is negative, $s_i$ is translated backward and while $\tau$ is positive, $s_i$ is translated forward, as illustrated in Fig. \ref{MSC}. $I(x,y)$ is the indicator function, equal to 1 if $x=y$ and 0 elsewhere.

After construction of NSG and SSG graphs, we adopt the graph matching technique, which is a powerful tool to solve mapping problems and has been widely used in computer vision and pattern recognition, to determine an optimal mapping between the two weighted graphs under the mapping rule. For these two graphs, we can compute their adjacency matrices, written as ${{A}_{n}}$ and ${{A}_{s}}$. The graph matching method is aimed to find out a permutation matrix $P$ that minimizes the following objective function:
\begin{equation}
	\underset{P}{\mathop{\min }}\,||{{A}_{n}}-P{{A}_{s}}{{P}^{T}}||_{F}^{2}
\label{ObjectFun}
\end{equation}
where $||\cdot |{{|}_{F}}$ denotes the Frobenius matrix norm. Solving this problem exactly is known to be an NP hard problem due to its combinatorial optimization property. Many algorithms have been proposed to find an approximated solution.

Among these algorithms the Factor Graph Matching (FGM) algorithm \cite{zhou2012factorized} has been demonstrated to produce state-of-art results. So here we utilize the FGM algorithm to solve the SSG to NSG matching problem in equation (\ref{ObjectFun}). Suppose in NSG the sum of graph edge weights of an vertex, say vertex ${{i}_{NSG}}\in {{V}_{NSG}}$, to all other vertices is $d({{i}_{NSG}})$, and, similarly, in SSG the sum of graph edge weights of vertex ${{i}_{SSG}}\in {{V}_{SSG}}$ to all other vertices is $c({{i}_{SSG}})$, then the difference between the normalized weight sum $\tilde{d}({{i}_{NSG}})$ and $\tilde{c}({{i}_{SSG}})$ reflects the similarity of the positions of ${{i}_{NSG}}$ and ${{i}_{SSG}}$ in the corresponding graphs, where $\tilde{d}({{i}_{NSG}})$ (and similarly $\tilde{c}({{i}_{SSG}})$) is the normalized weight sum of vertex ${{i}_{NSG}}$ that is obtained by dividing $d({{i}_{NSG}})$ by the largest weight sum in graph NSG. So we define the vertex similarity as:
\begin{equation}
\label{NodeSim}
	\exp \left( -|\tilde{d}({{i}_{NSG}})-\tilde{c}({{i}_{SSG}}){{|}^{2}}/2\sigma _{n}^{2}\  \right);\ \ {{i}_{NSG}},{{i}_{SSG}}=1...v
\end{equation}
and the edge similarity:
\begin{equation}
\label{EdgeSim}
	\exp \left( {-|a_{ij}^{NSG}-a_{kl}^{SSG}{{|}^{2}}}/{2\sigma _{e}^{2}}\; \right);\ \ i,j,k,l=1...v
\end{equation}
where: $a_{ij}^{NSG}$, $a_{kl}^{SSG}$ are graph edge weights in NSG and SSG, respectively; ${{\sigma }_{n}}$ and ${{\sigma }_{e}}$ are parameters to control the affinity between neurons and edges, respectively. Table \ref{OptMappingTable} gives the algorithm procedure.

\begin{table}[ht]
\caption{Graph matching based input mapping algorithm}\label{OptMappingTable}
\centering  
\renewcommand\arraystretch{0.4} 
\begin{tabular*}{7.5cm}{@{}cl}\toprule \midrule
  Step             &     \\ \midrule \midrule
1   & Input spike trains and input mapping neurons  \\
2 & Construct graphs NSG using Euclidean distance \\
3 & Construct graphs SSG using equation (\ref{SpikeSim1}) or (\ref{SpikeSim2}) \\
4 & Compute vertex similarity using equation (\ref{NodeSim}) \\
5 & Compute edge similarity using equation (\ref{EdgeSim})  \\
6  &Solve problem (\ref{ObjectFun}) by Factor Graph Matching \\ \midrule \bottomrule
\end{tabular*}
\end{table}

\begin{figure}[ht]
  \centering
  \includegraphics[width=6.5cm]{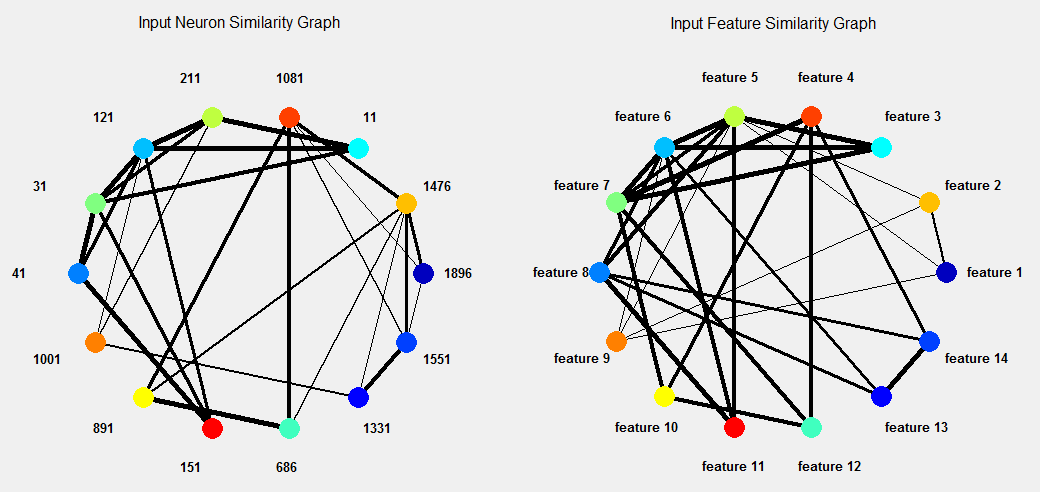}
  \caption{An input mapping result obtained by the proposed method for exemplar temporal data.}\label{InputMapping}
\end{figure}

Fig. \ref{InputMapping} shows matching results for an exemplar temporal data represented by 14 features (this is actually the case study data from experiment 1, presented in section 4 in \cite{tu2014neucube}). The left graph is the input NSG and the right graph is SSG. We can see that after matching, highly correlated features are mapped to nearby input neurons.

\subsection{The optimal input mapping enables early and accurate event prediction from temporal data}
\begin{figure}[ht]
  \centering
  \includegraphics[width=5.5cm]{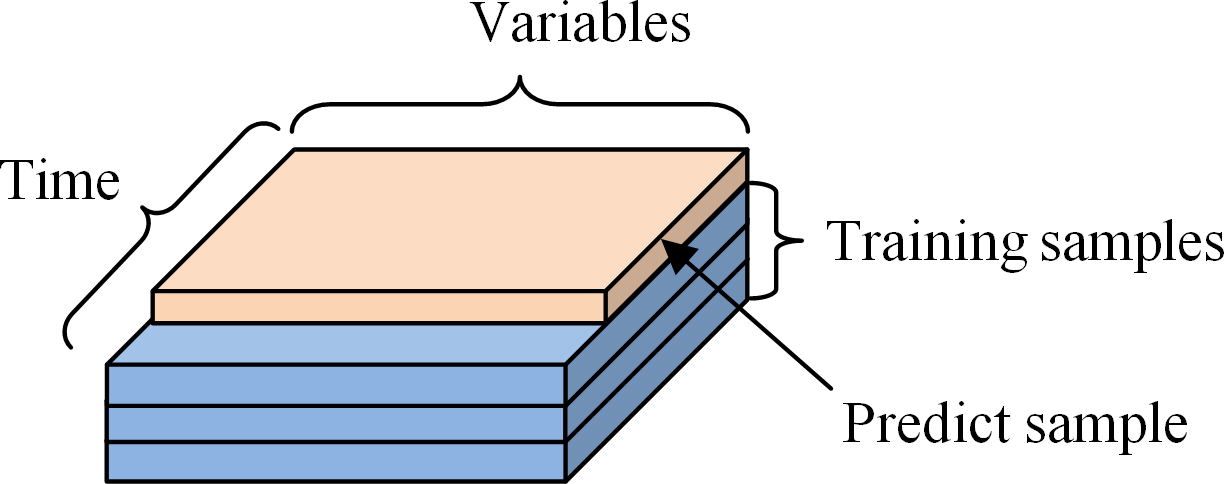}

  \caption{A temporal data model used for early event prediction}\label{datamodel}
\end{figure}

In many applications, such as pest population outbreak prevention, natural disaster warning and financial crisis forecasting, it is important to know the risk of event occurrence as early as possible in order to take some actions to prevent it or make adjustment in time, rather than waiting for the whole pattern of temporal data to be entered into the model. The main challenge in the early event prediction task is that the time length of the recall samples should be smaller than that of training samples, because training samples are collected in the past and all the data are already known. This  is illustrated in Fig. \ref{datamodel}.
Traditional data modeling methods, such as SVM, \emph{k}NN, and MLP, are no longer applicable for early event prediction task, because they require prediction sample and training samples to have same time length. Furthermore, traditional methods also cannot model both the close interaction and interrelationship between time and space components of the spatiotemporal data.

\begin{figure}[ht]
  \centering
   \subfigure[]{
  \includegraphics[width=2cm]{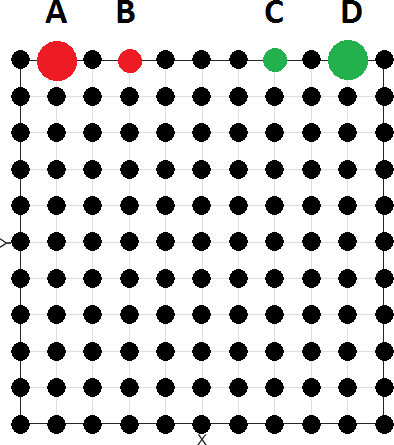}}
    \subfigure[]{
  \includegraphics[width=2cm]{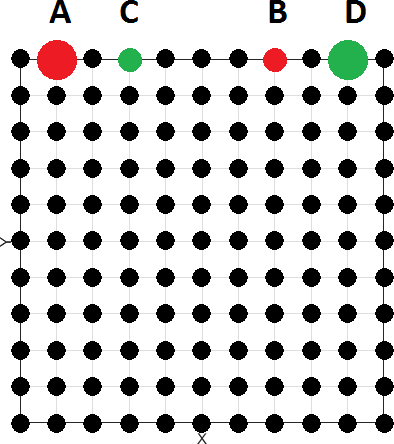}}
    \subfigure[]{
  \includegraphics[width=2cm]{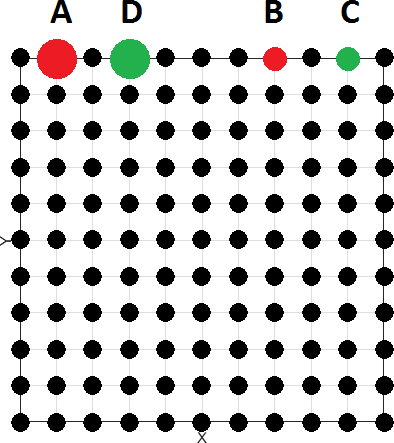}}
  \caption{Different spatial pattern of input mapping. Colored circles are input neurons and black dots are normal neurons. The color indicates correlation and shape size indicates dominance. }\label{MappingIllustration}
\end{figure}

In contrast, NeuCube with the proposed new mapping method would enable a better early event prediction, as the connectivity of a trained SNNcube would reflect on the temporal relationship in the temporal data. As a result, if part of a new sample is presented, this would fire a chain of activities in the SNNcube based on the established connections.  To be more specific, let us consider  mapping four features A, B, C and D into a two dimensional SNNcube\footnote{Recall that the size of a three dimensional SNNcube is $N=n_x\times n_y \times n_z$. Two dimensional SNNcube is a special case where $n_z=1$. }. Assuming that feature A is more correlated with feature B than with others, but feature A is a more dominant (important) feature than feature B, e.g. feature A could be Sun Radiation and  feature B could be temperature as in the Aphids study in \cite{tu2014neucube}. Similarly, assuming feature C is more correlated with feature D, but feature D is a more dominant feature than feature C. The result of optimal graph mapping are given in Fig. \ref{MappingIllustration} (a). The highly correlated features are mapped nearby while less correlated features are mapped far away. Recall that STDP learning rule adjusts synaptic weight between nearby neurons according to their firing time difference. During training process, the neurons around feature A will have more interaction with neurons around feature B than with other features, and these interactions cause unsymmetrical weight adjustment according to STDP rule. After training, the connection patterns between neurons around these two features will encode the temporal characteristics in the training spike trains, hence the  temporal relationship in original signals.

During testing phase, once similar temporal pattern appears, early firing is triggered and then propagated in a neuron population (hence a firing chain of a neuron cluster) before full data are presented, i.e. the neurons and their connections form a functional cluster which becomes selective to the spatiotemporal pattern and tend to fire at the start of the signal pattern. As demonstrated by \cite{masquelier2008spike,masquelier2009competitive},  LIF neurons equipped with STDP learning rule can learn unsupervisedly an arbitrary spatiotemporal pattern embedded in complex background spike trains and when its preferred spike sequence appears, the neuron can emit a spike very early at the start of the pattern.  Biologically, the population firing in the SNNcube chain-fire phenomenon was observed in zebra finches HVC area to control the precise temporal structure in birdsong \cite{ikegaya2004synfire}, in which the neural activity is propagated in chain network to form the basic clock of the song rhythm. In the SNNcube we have also observed a similar chain-fire phenomenon while spike trains are presented to the network. These features endow the SNNcube with a powerful potential to encode complex spatiotemporal patterns contained in the input spike trains used for training and to respond early to the presence of a specific spatiotemporal pattern in a recall/prediction mode.  This is especially important when a result is the consequence of   several highly correlated key factors, as further demonstrated by the study of Stroke occurrence \cite{othman2015spatial}\footnote{http://www.kedri.aut.ac.nz/neucube/stroke}. This phenomenon is similar to that of associative memories in Hopfield networks \cite{hopfield1982neural}, but here we deal with temporal patterns rather than with static input vectors.

In contrast,  Fig. \ref{MappingIllustration} (b) and (c) display the non-optimal mapping, where uncorrelated features are mapped together. The neurons around uncorrelated features can hardly learn the original signal temporal relationship, because nearby neurons are presented with less even non-correlated signals.  The interactions between these nearby neurons can be much lower and do not capture any meaningful temporal correlation. As a result, while new samples are partially presented during testing, similar temporal pattern as contained in training samples cannot be detected correctly, as will be demonstrated by experimental results in Section IV.

\subsection{The optimal input mapping enables better network structure analysis and visualization and a better data understanding}
After it is trained, the SNNcube has captured spatial and temporal relationships from the temporal data. It is helpful to know how the neurons in the SNNcube are related to the input features and what patterns  the SNNcube have learned from the input signals. This information is important to understand the model and the temporal data set \cite{capecci2015analysis}.  While in previous work the neuronal clusters are manually labeled according to the synaptic weights and this costs plenty of time and is less accurate, we propose the following algorithm to  unveil SNNcube structure through automatically analysing neuronal clusters in the SNNcube.

The neurons in the cube are indexed from 1 to $N$ according to  ascendent order of their $x$, $y$ and $z$ coordinates. We mark the input neurons as the information source in the SNNcube and define a source matrix ${{F}_{src}}\in {{R}^{N\times v}}$ as follows: if neuron $i$ is the input neuron of variable $j$, then the entry ($i$, $j$) of ${{F}_{src}}$ is 1, otherwise is 0. The affinity matrix $A\in {{R}^{N\times N}}$ of the SNNcube is defined in the following way: the entry ($i$, $j$) of $A$ is the total spike amount transmitted between neuron $i$ and neuron $j$. Note that more spike means stronger interaction and more information shared between the neurons. Then the ratio of information for each neuron received from the input information sources is computed by the following algorithm:
\begin{enumerate}
  \item  Compute $S={{D}^{-1/2}}A{{D}^{-1/2}}$
  \item Evaluate $\tilde{F}={{I}_{rate}}S\tilde{F}+(I-{{I}_{rate}}){{F}_{src}}$ to converge
  \item Normalize $F={{G}^{-1}}\tilde{F}$
\end{enumerate}

$I$ is the identity matrix and ${{I}_{rate}}$ is a diagonal matrix defining the propagation rates on different directions. $D$ and $G$ are also diagonal matrices with ${{D}_{ii}}=\sum\nolimits_{j=1}^{N}{{{A}_{ij}}}$ and  ${{G}_{ii}}=\sum\nolimits_{j=1}^{N}{{{{\tilde{F}}}_{ij}}}, i=1,2,...,N$, respectively. In the first iteration $\tilde{F}={{F}_{src}}$. Step 1 computes the normalized adjacency matrix which can fully encode the connection information of SNNcube by a square matrix, according to network theory \cite{chung1997spectral}. Step 2 propagates information from input neurons to other neurons in an iterative way. The main principle behind the algorithm is that information (or activation) is propagated in the network and the propagation process is dominated by the network structure. Imagining that each input neuron is a source and has some kind of information (i.e. electricity, fluid matters etc.),  but the information type possessed by different input neuron is different (i.e. different ions or different liquid). The information is propagated from sources to other neurons in each iteration and the propagation amount is proportional to the connection weight between each pair of neurons. At the beginning only the input neurons (the sources) have the information and other neurons don't have any information. As the propagation process continues, all neurons receive some information from one or more input neurons and the information amount corresponding to different input neurons is different. The amount of a particular type of information (from the corresponding input neuron) received by a neuron reflects the intimation relationship of this neuron with that input source neuron. The more amount of information it received, the closer it is with that input neuron. Finally according to network theory when the whole network reaches to a stable state,  entry $F_{ij}$ represents the relative information amount neuron $i$ received from input neuron $j$ \cite{shrager1987observation, zhou2004learning}.  Finally step 3 normalizes the information amount corresponding to different input neurons and this facilitates to classify neurons into different input clusters by $k=\arg \underset{i=1..v}{\mathop{\max }}\,{{F}_{ij}},j=1...N$, where $v$ is the number of input neuron and $N$ is the number of total neurons in SNNcube. Matrix $F$ is the basis of extracting neuronal clusters.


The propagation factor matrix ${{I}_{rate}}$ controls the convergence of the propagation algorithm and the amount of the information being propagated to other neurons in the SNNcube. Here ${{I}_{rate}}$ is computed by ${{({{I}_{rate}})}_{ii}}=\exp \left( -{{{\bar{d}}}^{2}}_{i}/2{{\sigma }^{2}}\  \right)$, where $\bar{d}_{i}^{{}}$ is the mean affinity value between a neuron and its 8 neighboring neurons, so that the information propagation between strongly connected neurons is large while the information propagated through weakly connected neurons is small.

We now show the propagation algorithm converges. Because $ I_{rate}$ is a diagonal matrix and ${{({{I}_{rate}})}_{ii}}=\exp \left( -{{{\bar{d}}}^{2}}_{i}/2{{\sigma }^{2}}\  \right)<1$, the spectral radius of $ I_{rate}$ is $\rho \left( {{I}_{rate}} \right)<1$. After the $i_{th}$ iteration, we have
\begin{equation}
\tilde{F}^{(i)} = A_{sc} F_{src} + A_{nb}(I-I_{rate})F_{src}
\end{equation}
where ${{A}_{sc}}={{({{I}_{rate}}S)}^{i-1}}$ and ${{A}_{nb}}=\sum\nolimits_{k=0}^{i}{{{({{I}_{rate}}S)}^{k}}}$, representing the information amount acquired from sources and neighborhood, respectively. Let $P={{D}^{-1}}A$ be the random walk matrix on the graph. Then for the spectral radius of $P$  we have $\rho (P)\le \|P\|_{\infty}\le 1$. Because $S={{D}^{1/2}}P{{D}^{-1/2}}$, $S$ is similar to $P$, so $\rho (S)=\rho (P)\le 1$.  Since the spectral radius $\rho ({{I}_{rate}})<1$, and $\rho (AB)\le \rho (A)\rho (B)$, so $\rho ({{I}_{rate}}S)<1$. Therefore $\underset{i\to \infty }{\mathop{\lim }}\,{{A}_{sc}}=0$, $\underset{i\to \infty }{\mathop{\lim }}\,{{A}_{nb}}={{(I-{{I}_{rate}}S)}^{-1}}$ and
\begin{equation}
{{F}^{*}}=\underset{i\to \infty }{\mathop{\lim }}\,{{\tilde{F}}^{(i)}}={{(I-{{I}_{rate}}S)}^{-1}}(I-{{I}_{rate}}){{F}_{src}}
\label{eqoptimal}
\end{equation}
It is worth mentioning that Step 2 can also be replaced by equation (\ref{eqoptimal}). But since the matrix $A$ is highly sparse, it will be much more efficient to evaluate the equation in Step 2 than to invert a matrix in equation (\ref{eqoptimal}), regarding to both space and time comsumption\footnote{ Note that ${I}_{rate}S$ is a highly sparse matrix which keeps constant during iterations and Step 2 usually converges after several iterations.}. Another advantage of using Step 2 is that during iterative process, it is more interesting and intuitive to observe how the information is propagated in the SNNcube, rather than  jumping directly to the resultant clusters view given by equation (\ref{eqoptimal}).

\begin{figure}[ht]
  \centering
  \includegraphics[width=5.3cm]{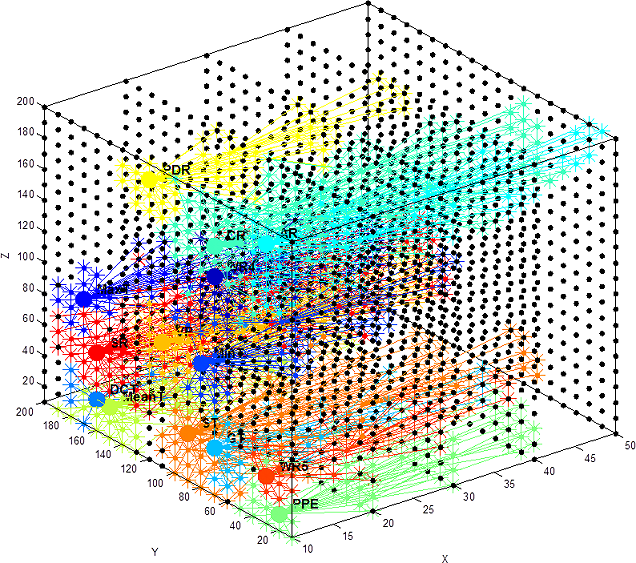}
  \includegraphics[width=3cm]{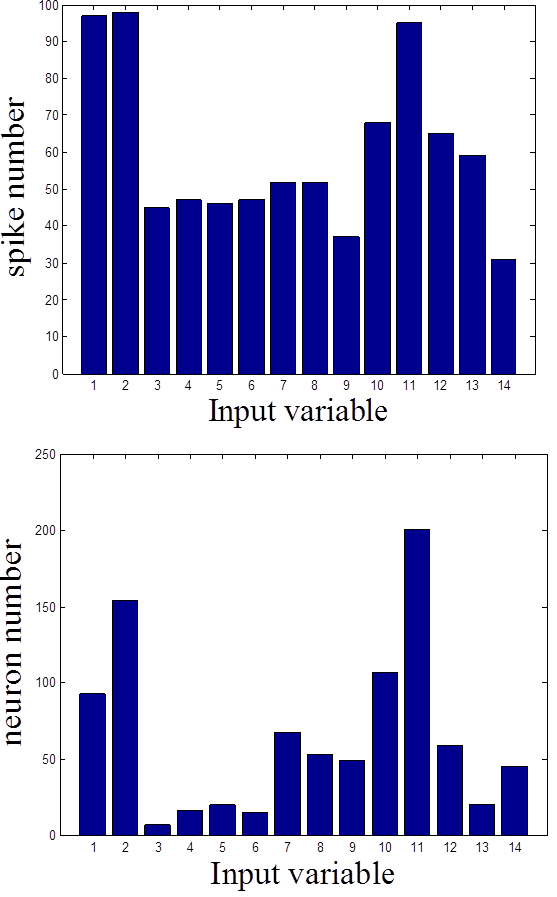}
  \caption{Left: Neuronal clusters in  SNNcube  after unsupervised training; Right: input spike number of each feature (top) and neuronal connections of each input neuron (bottom).}\label{NeuCubeVisualization}
\end{figure}

Fig. \ref{NeuCubeVisualization}  left plot shows the network structure after unsupervised training for the study of Aphids data presented in our previous work \cite{tu2014neucube}. The big solid dots represent input neurons and other neurons are labeled in the same intensity as the input neuron from which it receives most spikes. The unconnected dots mean no spike arrived at that neuron. On Fig. \ref{NeuCubeVisualization} right, the top pane is spike number for each variable after encoding and the bottom pane is the neuron number belonging to each input variable cluster. From this figure we can see the consistency between the input signal spike train and the SNNcube structure. Note that variable 11 (solar radiation) is emphasized in the SNNCube that suggests a greater impact of the solar radiation on the aphid number. This was observed also in a other work \cite{worner2002improving}. This is very different from traditional methods such SVM which have been used for same tasks but without offering facilities to trace the learning processes for the sake of data understanding. It is worth mentioning that the spatial pattern of the input mapping, e.g. in Fig. \ref{InputMapping}, is embedded in the source matrix $F_{src}$  and has a direct influence on the visualization results and the interpretation of the results. These mapping and visualization are demonstrated to be useful for high level \emph{cause - results} interpretation in some studied, such as EEG signal study for identifying differences between people with opiate addiction and those undertaking substitution treatment for opiate addiction \cite{capecci2015analysis} and the  fMRI data study for brain activity while subjects are presented with different pictures\footnote{http://www.kedri.aut.ac.nz/neucube/fmri.}. It is important to note that while in \cite{capecci2015analysis} the input mapping and neuronal clustering were performed by hand, the proposed graph mapping and structure analysis algorithm enable all these to be finished automatically.
\section{Experimental results on three case study problems }
In this section we present three case studies to demonstrate the validity of the proposed architecture for both early even prediction based on temporal data and spatiotemporal pattern recognition. The first case study is conducted on a benchmark physiological data set SantaFe{\footnote{http://www.physionet.org/physiobank/database/santa-fe/}} to demonstrate the ability of the proposed method to classification temporal data and to make early event prediction. We demonstrate the validity of the mapping method for early event prediction. The second case study is conducted on a spatiotemporal data set PEMS, which can be downloaded from California Department of Transportation PEMS website\footnote{http://pems.dot.ca.gov}, to perform pattern recognition, in which each feature is sampled from a fixed location. We demonstrate that the proposed mapping in section 3 improves the accuracy of  spatiotemporal pattern recognition. The third case study is carried on a contest physiological data set Challenge 2012\footnote{http://physionet.org/challenge/2012/\#rules-and-dates} to demonstrate the superiority of the proposed graph mapping over randomly mapping .

While in \cite{kasabov2014neucube, kasabov2012neucube, kasabov2014evolving} address event representation (AER) encoding with fixed threshold is used and the threshold has to be tuned manually each time for each feature individually,  here  we introduce a self-adaptive bi-directional thresholding method, Adaptive Threshold Based (ATB)  encoding algorithm. ATB is simply self-adaptive to all features: for an input time series/signal $f(t)$, we calculate the mean $m$ and standard deviation $s$ of the gradient $df/dt$, then the threshold is set to $m+\alpha s$, where $\alpha $ is a parameter controlling the spiking rate after encoding. After this, we obtain a ‘positive’ spike train which encodes the segments in the time series with increasing signal and a negative spike train, which encodes the segments of decreasing signal.



Because NeuCube system is a complex system and contains many tunable parameters, manually tuning these parameters might be time consumption for a non-experienced user. We have implemented a Genetic Algorithm (GA) optimization component to optimize all the system parameters at once in a $k$-fold cross validation way with respect to a data set. The objective function of the GA is the overall validation error rate. We use the optimization component to simultaneously optimize 7 key parameters of the NeuCube system in Table \ref{OptParamTable}. After optimization, GA outputs the optimal parameters' value that attained the smallest cross validation error rate.
\begin{table}[ht]
\caption{Parameters to be included in the optimization process}\label{OptParamTable}
\centering  
\renewcommand\arraystretch{0.4} 
\scriptsize
\begin{tabular*}{8.5cm}{@{}ccc}\toprule \midrule
  Parameter             &Meaning      & Value range\\ \midrule \midrule
Spike Threshold   & Threshold of ATB encoding  & [0.1, 0.9]\\
Firing Threshold & LIF neuron firing threshold & [0.01, 0.8]\\
STDP Rate & The learning rate in STDP algorithm & [0.001, 0.5]\\
Refractory Time &Refractory time after firing &[2, 9]\\
Mod &Modulation factor in \cite{kasabov2013dynamic} &[0.00001,0.5]\\
Drift & Synaptic drift parameter in \cite{kasabov2013dynamic}&[0.1, 0.95]\\
$K$ & Number of nearest neighbors in deSNN &  [1, 10] \\ \midrule \bottomrule
\end{tabular*}
\end{table}

Baseline algorithms consist of: Multiply Linear Regression (MLR) is simple, extensively studied and can achieve relative stable result in many applications; Support Vector Machine (SVM) is probably the most widely used and has been demonstrated to be successful in various applications; Multilayer Perceptron (MLP) is a classic neural network model and has advantages in processing multivariate data; $k$ Nearest Neighbors ($k$NN) and weighted $k$ Nearest Neighbors (w$k$NN) \cite{hechenbichler2004weighted} are among the most popular classification algorithms \cite{wu2008top} and especially while processing high-dimension data; finally, Global Alignment Kernel (GAK) \cite{cuturi2011fast} is a recently developed method to process time series and can achieve the state-of-art results in many applications. So there are 6 baseline algorithms\footnote{The first three methods are implemented in NeuCom system: http://www.kedri.aut.ac.nz/areas-of-expertise/data-mining-and-decision-support-systems/neucom. $k$NN and w$k$NN are implemented based on MATLAB \emph{knnsearch} function. GAK is available at: http://www.iip.ist.i.kyoto-u.ac.jp/member/cuturi/GA.html}: MLR, SVM, MLP, $k$NN, w$k$NN and GAK.

\begin{figure}[ht]
  \centering
  \includegraphics[width=8cm]{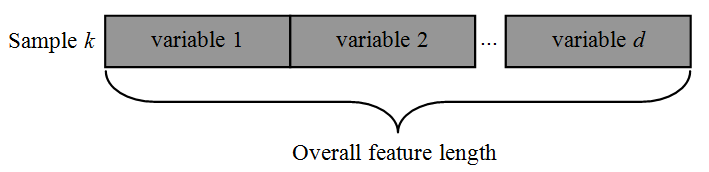}
  \caption{Convert  spatiotemporal data to static vector for baseline algorithms. }\label{FeatureForm}
\end{figure}

Because baseline algorithms process static vectors, we prepared their data set in the following way. We concatenated each temporal variables one after another to form a long feature vector for each sample, as shown in Fig. \ref{FeatureForm}.
Since baseline algorithms have very few tunable parameters ( one or two parameters), the parameters are tuned in a grid search method. This process is simple and quick, usually just a few steps to locate the optimal value of the parameters.

\subsection{Predict events by classifying time series data based on physiologic signals}
In this case study we conduct experiments for complex physiologic time series classification on SantaFe dataset. The dataset contains three physiologic features: heart rate (HR), respiration force (RF, chest volume) and blood oxygen (BO). Each of the feature was measured twice in one second and the measurement lasted 5 hours. The aim of this case study is to predict the sleep stage of a patient by using the collected physiologic signals. There are three sleeping stages: awake, sleep and waking, and the sleep stages are read from EEG signals by neurologist.

The original dataset contains 34000 data points for each feature (approximate 4.7 hours long) and 103 sleep stage labels, and the sleep stage labeling interval varies from 60 seconds to 4600 seconds. Each label indicates a different sleep stage happening at that moment, i.e. whenever the patient transits from one sleep stage to another different sleep stage, a corresponding label is assigned at that time. So the signals recorded during the labeling interval reflect the happening process of the next sleep stage. Consequently by classifying these temporal data into different classes, we actually predict the next sleep stage of the patient. In this case study we only use the labeling data with 60 seconds time interval to make sure the feature length is same for every sample for fair comparison, because traditional methods (such as SVM) cannot process data which have uneven feature length between samples.  In this case we use a $10\times 10\times 10$ SNNcube\footnote{Currently we determine the reservoir size by trying several different ones. But as a general rule, the larger the reservoir is, the more powerful computational ability the reservoir has and thus the more complex patterns the model can learn and recognize. However, a larger reservoir consumes longer time and more memory space and needs more training data for training.} and deSNN network with a weighted $k$NN classifier as the output layer \cite{kasabov2013dynamic}. We build a $60\times3\times42$ data matrix, where 60 is the time length of each signal, and 3 is the feature number and 42 is the sample number.

\begin{figure}[ht]
  \centering
  \includegraphics[width=4cm]{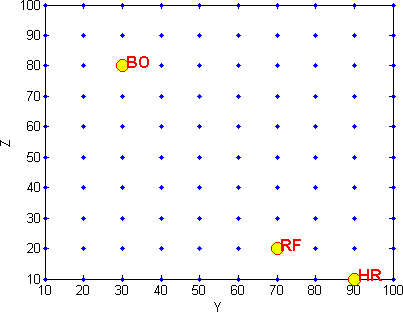}
  \includegraphics[width=4cm]{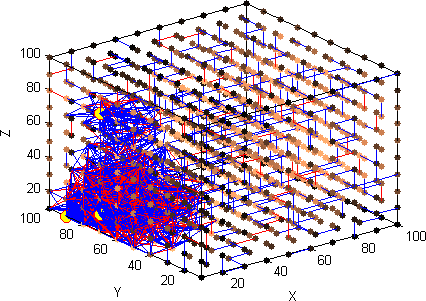}
  \caption{Left: input Mapping for SantaFe dataset (Y-Z plane of the SNNcube). Right: synaptic connections after training (Blue connections have positive weight and red connections have negative weight).}\label{SantaFeMapping}
\end{figure}

Fig. \ref{SantaFeMapping} left pane gives the mapping result found by the proposed input mapping algorithm. It can be seen that the respiration force is mapped between blood oxygen and heart rate, but it is much closer to heart rate. This indicates the respiration signal more correlates with heart rate than blood oxygen. This result is consistent with the actual observation described in the three major research questions, which are raised on the dataset website \footnote{How do the evolution of the heart rate, the respiration rate, and the blood oxygen concentration affect each other? (A correlation between breathing and the heart rate, called respiratory sinus arrhythmia, is almost always observed.)}. This is also further demonstrated by Fig. \ref{SantaFeMapping} right plot, in which we can see clearly that the connections between respiration force and heart rate are much denser than connections in other places. This indicates that the interaction between respiration force and heart rate is more active and much stronger than other parts. Therefore, the temporal relationship in the original data is fully captured by the connections in the SNNCube after training.

\begin{figure}[ht]
  \centering
  \subfigure[]{
  \includegraphics[width=4cm]{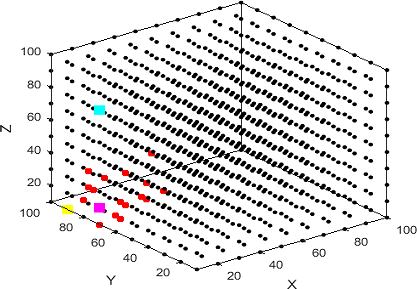}}
  \subfigure[]{
  \includegraphics[width=4cm]{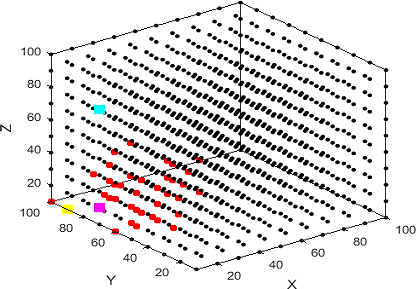}}

  \subfigure[]{
  \includegraphics[width=4cm]{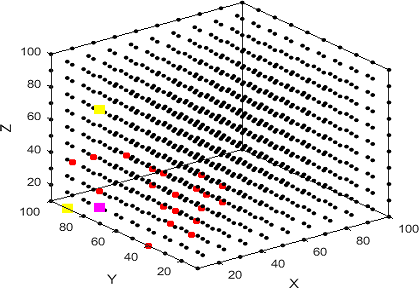}}
  \subfigure[]{
  \includegraphics[width=4cm]{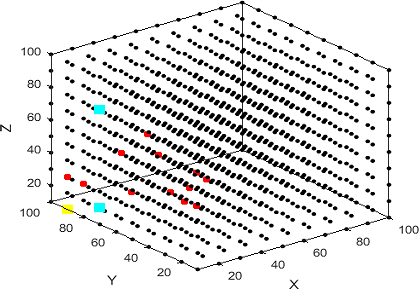}}
  \caption{Firing state spreading in the SNNcube during training stage. From (a) to (d) four consecutive frames are displayed. }\label{NeuronFiringState}
\end{figure}

Fig. \ref{NeuronFiringState} shows four consecutive snapshots of the instantaneous firing state of the neurons during training process. Squares are input neurons (Purple means that a positive spike is input at this moment. Cyan means negative spike and yellow means no spike). Red dots are firing neurons and black dots are unfiring neurons. Note that the firing neuronal cluster grows from small to large and the propagation of the spike states (hence the information the spike carried) has a trajectory in the SNNcube. While the meaning of the firing pattern and spreading trajectory, how they reveal the significance of the SNNcube modelling ability are questions to be answered, the study of these is out of the scope of this paper and we leave them in our future papers.

\begin{figure}[ht]
  \centering
  \subfigure[]{
  \includegraphics[width=4cm]{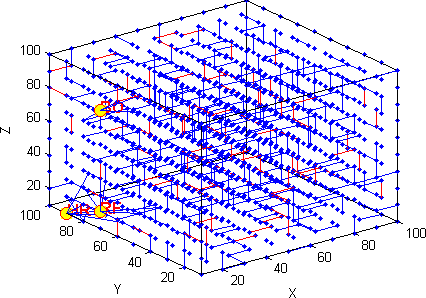}}
  \subfigure[]{
  \includegraphics[width=4cm]{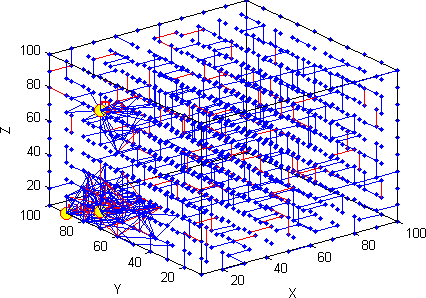}}

  \subfigure[]{
  \includegraphics[width=4cm]{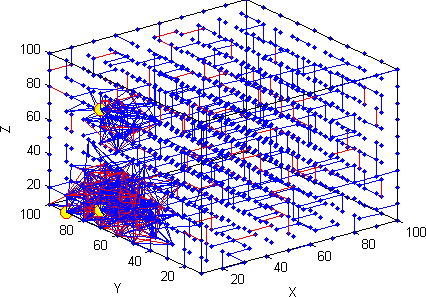}}
  \subfigure[]{
  \includegraphics[width=4cm]{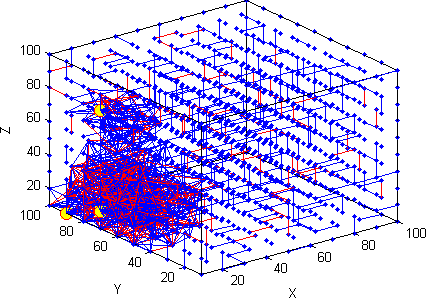}}
  \caption{Connection evolving process during training stage.}\label{ConnectionEvolving}
\end{figure}

\begin{figure}[ht]
  \centering
  \includegraphics[width=7cm]{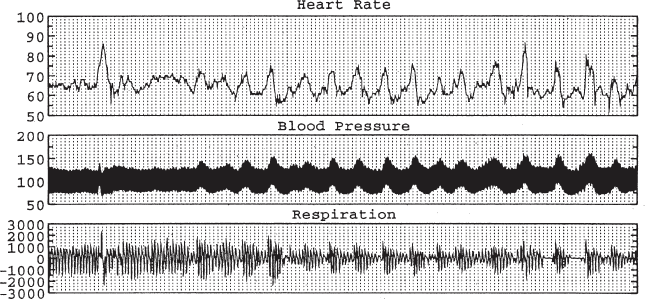}

  \caption{The physiological signals in SantaFe dataset (Courtesy from \cite{ichimaru1999development}).}\label{PhysioSignals}
\end{figure}

 Figure \ref{ConnectionEvolving} shows the evolving process of the connections during different time in the training stage. From this we can see how the connections between different neurons are created dynamically as the spike trains input to the reservoir. At the beginning, the connections in the reservoir are sparse and nearly randomly distributed \ref{ConnectionEvolving}(a). As the training process continues, some new connections appear around each input neuron, but blood oxygen input neuron does not have interaction with the other two input neurons \ref{ConnectionEvolving}(b). At the last half training process, more connections are created and the interaction between each pair of input neurons increases, especially for blood oxygen \ref{ConnectionEvolving}(c)-(d). The synaptic weight evolving process is tightly connected with the physiological signal trend in Fig. \ref{PhysioSignals}, from which we can see clearly the correlation between blood oxygen and the other two signals in the last half signal segment. It is worth mentioning that while previous method to study the SantaFe dataset mainly focused on analysis of the nonlinear dynamics of the respiration, heart rate and blood pressure \cite{korurek2010clustering, engin2004ecg}, they can hardly provide such an intuitive and direct way to observe the dynamics of the signals, and hence the interaction between the physiological signals during a sleep stage.

We designed two experiments on this data set to show the predictive ability of NeuCube with different mapping and how early the model can predict sleep stage using part of the data. In these experiments, we trained NeuCubes of both random mapping and graph mapping using 100\% of the time length samples (60 seconds), but temporal data of only 75\% and 50\% of the time length of the samples was used to predict the sleep stage.
%

Experimental results of early event prediction are in Table \ref{AccuracyTable2}. In order to comparing the impact of testing time length upon accuracy, we also include the result of full time length in the second column. NeuCube(R) is random mapping and NeuCube(G) is graph mapping.
\begin{table}[ht]
\caption{Early Prediction Accuracy of SantaFe Dataset (\%)}\label{AccuracyTable2}
\centering  
\renewcommand\arraystretch{0.4}
\scriptsize
\begin{tabular*}{8.5cm}{@{\extracolsep{\fill}}cccccccc}\toprule \midrule
 & \multicolumn{7}{c}{Accuracy of each training and testing time length (sec)}                  \\ \cmidrule(r){2-8}
                             &&& 60                    &&& 45                 &30                   \\
                             &&& (full)               &&&(early)               &                     \\ \midrule \midrule
MLR                    &&& 42.50              &&& 32.29               & 32.29               \\ \cmidrule(r){2-8}
SVM                    &&& 43.75              &&& 22.29               & 28.96               \\ \cmidrule(r){2-8}
MLP                    &&& 38.96              &&& 39.17              & 26.25      \\ \cmidrule(r){2-8}
\emph{k}NN      &&& 35.85               &&& 38.75               & 32.08               \\ \cmidrule(r){2-8}
w\emph{k}NN   &&& 48.45               &&& 48.13               & 25.83               \\ \cmidrule(r){2-8}
GAK             &&& 45.21                &&& 40.07               & 38.43               \\ \cmidrule(r){1-8}
NeuCube(R)   &&& 51.62               &&& 43.5               & 35.12               \\ \cmidrule(r){1-8}
NeuCube(G)              &&& \textbf{67.74}  &&& \textbf{63.06}                & \textbf{51.61}               \\ \midrule \bottomrule
\end{tabular*}

\end{table}

From these results we can see that NeuCube(G) can perform better for early event prediction. In contrast the performance of NeuCube(R) for early prediction is much lower, because SNNcube with randomly mapping fails to acquire the temporal pattern  in the data for early firing during testing stage. Furthermore, a realistic early event prediction should be that as the time length of observed data increases, the prediction accuracy will also increase. But from table \ref{AccuracyTable2} we can see that as the time length of training data increases, traditional modeling methods do not necessarily produce better results (some even worsen), because they cannot model the whole spatiotemporal relationship in the prediction task. They can only model a certain time \emph{segment}. Temporal data typically exhibits complex interrelationship and interaction among different features and traditional methods are proposed to process static vector data and thus they can hardly model the complex relationship contained in temporal data. The GAK and w$k$NN can produce better results, but their accuracies are still much lower than that of NeuCube. Because NeuCube is trained on the whole spatiotemporal relationship in the data, even a small amount of input data can trigger spiking activities in the SNNcube that will correspond to the learned full temporal pattern resulting in a better prediction.

\subsection{Spatiotemporal pattern recognition based on spatiotemporal data }
In this case study we consider a benchmark traffic status classification problem of spatiotemporal data from the PEMS database. In freeways, vehicle flow is monitored by traffic sensors with fixed spatial locations and the data collected by these sensors exhibit spatial and temporal characteristics. Discovering spatial-temporal patterns can be very meaningful for traffic management and a city traffic plan.

\begin{figure}[ht]
  \centering
  \includegraphics[width=3cm]{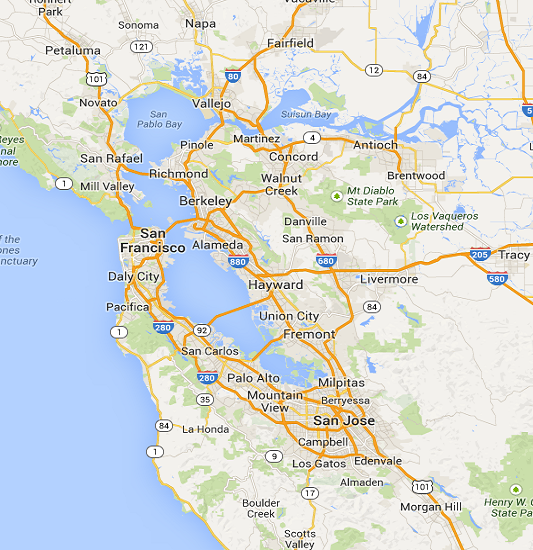}
  \includegraphics[width=3cm]{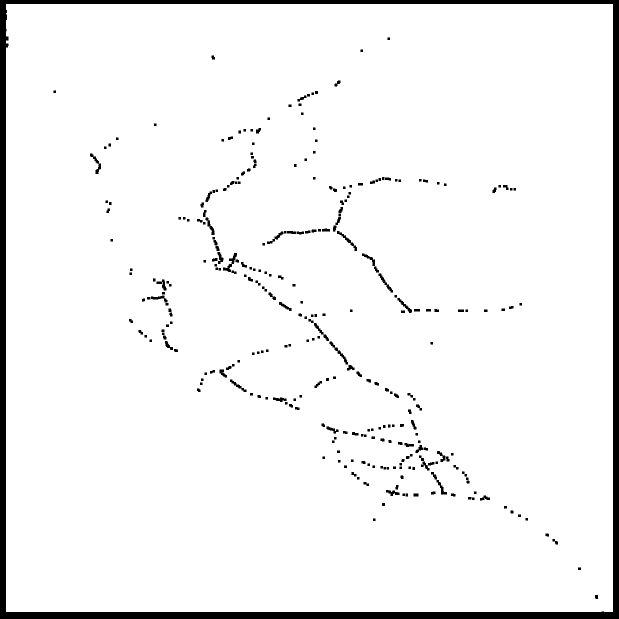}

  \caption{Left: map of the study area (From Google map).  Right: a reconstructed topology of the traffic sensor network (Each black dot inside represents a sensor.}\label{StudyAreaMap}
\end{figure}

\begin{figure}[ht]
  \centering
  \includegraphics[width=9cm]{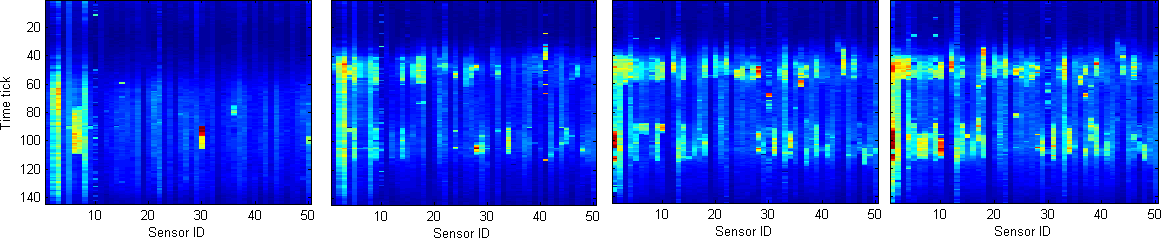}

%
  \caption{The spatial and temporal patterns of the sample traffic data from Monday to Thursday}\label{TrafficDataPattern}
\end{figure}

The study area is San Francisco bay area which is shown in Fig. \ref{StudyAreaMap}. There are thousands of sensors distributed over the road network and the sensors distribution is indicated in right plot of Fig \ref{StudyAreaMap}, in which each black dot represents a monitoring sensor. These sensors monitor lane occupation rate 24-hourly every day. Measurements are taken every 10 minutes and normalized between [0 1], where 0 means no car occupation and 1 means full occupation of the lane in the monitoring region. So there are 144 (24x6) data points per day. In this case study we collect traffic data over a period of 15 months and thus, after removing public holidays and sensor maintenance days, there are 440 days to be classified.
We did some preprocessing of the data: (1) we removed the data of outlier sensors from the data set, e.g. sensors producing always 1 or 0 in 24 hours and sensors flip from 0 to 1 or 1 to 0 suddenly; (2) nearby sensors that produce almost the same data sequence are combined into one sensor; (3) the total occupation rate of each sensor is calculated as a sum of all measurements over 440 days; (4) 50 sensors corresponding to the largest occupation rate are selected as the final features (variables) to represent the data set. Fig. \ref{TrafficDataPattern} shows samples of the spatial and temporal distribution of the traffic status of the road network from Monday to Thursday.

\begin{figure}[ht]
  \centering
  \includegraphics[width=6cm]{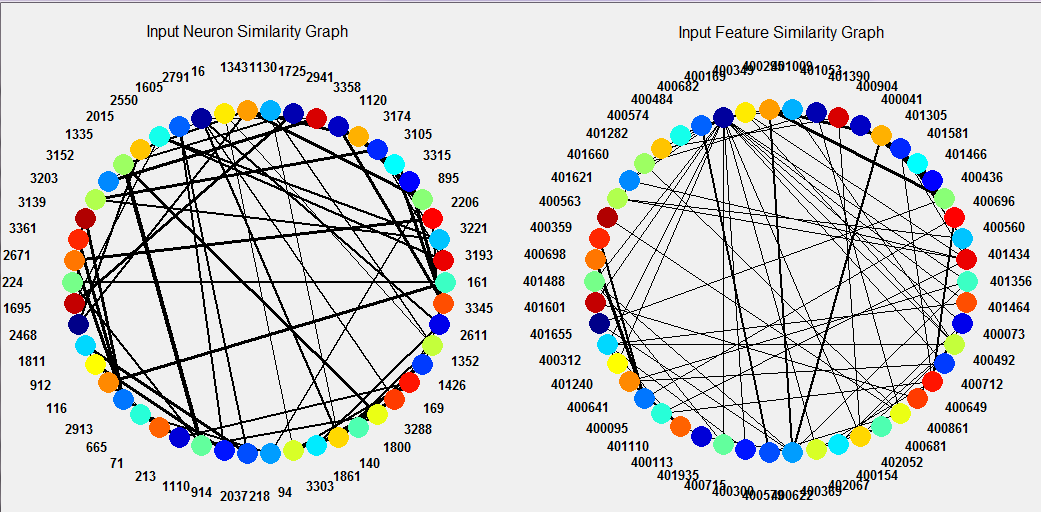}

  \caption{The final input mapping result used in this case study. Left: input neuron similarity graph (the number beside each vertex is the input neuron ID); right: input feature similarity graph  (the number beside each vertex is the traffic sensor ID).}\label{PEMSFeatureMatch}
\end{figure}

In this case study we construct a $15 \times 15 \times 15$ reservoir. Fig. \ref{PEMSFeatureMatch} shows the final input mapping result found by the graph matching algorithm. We can see the similar connection pattern between the two graphs.
\begin{figure}[ht]
  \centering

  \includegraphics[width=8cm]{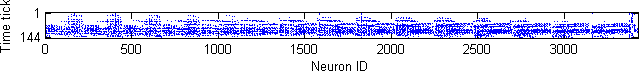}
   \includegraphics[width=8cm]{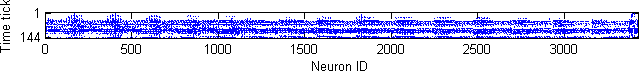}
    \includegraphics[width=8cm]{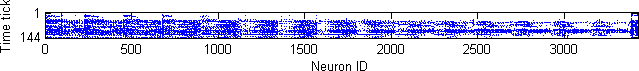}
     \includegraphics[width=8cm]{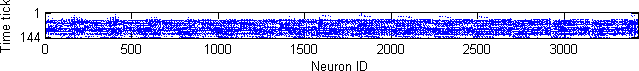}

  \caption{Neuron firing state of the reservoir. Bars from top to bottom correspond to Monday to Sunday.}\label{SpkPattern}
\end{figure}

Fig. \ref{SpkPattern} shows the overall neuron firing state matrix of the SNNcube corresponding to the four data samples, from top to bottom Monday till Thursday. In each figure, the horizontal axis is neuron ID and the vertical axis is time tick, from 0 at the top to 144 at the bottom. One should note that while in the plot it seems the firing state matrix is very dense, it is actually very sparse. Take the Thursday (the fourth one) as an example. There are 20416 firing entries and the firing state matrix size is 486000 ($144\times 3375$, where 3375 is the total neuron number in the SNNcube), so the firing entries is about 4.20\%. We can see that these sparse firing matrices have different patterns related to the input data. Meanwhile, since the size of the SNNcube can be specified according to the problem, the SNNcube with highly sparse firing rate has a great power to encode input signals and patterns, and thus it can potentially model any complex spatial and temporal relationship.

After obtaining the firing state matrix of each sample, we transmit them to the output deSNN layer \cite{kasabov2013dynamic}, which creates an output neuron for each sample and connects the output neuron to each neuron in the SNNcube. Then the weights of the connections are  established by Rank Order (RO) learning rule \cite{thorpe1998rank}. For a testing sample, its class label is determined by comparing its output neuron weights with those training samples' output neuron weights, which are established at training stage, using the weighted $k$NN rule \cite{hechenbichler2004weighted}. In Table \ref{AccuracyTablePEMS}, we compared the 2-fold cross validation experimental result between NeuCube and baseline algorithms: MLR, SVM, MLP, $k$NN, w$k$NN and Global Alignment Kernels (GAK) \cite{cuturi2011fast}. 
%

\begin{table}[ht]
\caption{Comparative accuracy of spatiotemporal pattern classification (\%)}\label{AccuracyTablePEMS}
\centering  
\renewcommand\arraystretch{0.4} 
\scriptsize
\begin{tabular*}{8.8cm}{@{}cccccccc}\toprule \midrule
               &MLR     &SVM     &MLP        &$k$NN     &w$k$NN     &GAK             &NeuCube  \\ \midrule \midrule
Param.   & -            &$d$:2   &$n$:100  &$k$:10   &$k$:10        &$\sigma$:5 & -                                  \\ \cmidrule(r){2-8}
Acc.       &56.82     &43.86    & 68.18      &66.82       &71.36           &72.27             &75.23                            \\ \midrule \bottomrule
\end{tabular*}

\end{table}

The parameters' values used in the classical machine learning methods are: $d$- degree of polynomial kernel; $n$ - number of neurons in the hidden layer of MLP; $k$-  number of nearest neighbors in $k$NN and w$k$NN;  $\sigma$-  Gaussian kernel width. From these results we can see that the proposed NeuCube model achieves better classification results. This is because traditional machine learning methods are designed to process static vector data, and they have limited ability to model spatially correlated and temporally varied data. Meanwhile, MLR, SVM and MLP also show disadvantages while modeling high-dimensionql data (e.g. there are 7200 features in each sample of this case study). $k$NN and w$k$NN have been widely used in high-dimension data processing, such as document classification, because they can approximately reconstruct the underlying manifold whose dimension is usually much lower than its ambient space and thus they can produce better results than MLR and SVM. While the recently proposed GAK algorithm is shown to be very efficient and effective in processing time series, its performance is still lower than that of NeuCube. It should be mentioned that although in this case study classifying weekday traffic patterns does not have an obvious application situation now,  this PEMS database has been established as a benchmark data set to test an algorithm's ability for spatiotemporal data processing \cite{cuturi2011fast, skabardonis2003measuring}.

\subsection{Comparison of random input mapping and graph input mapping method for Physionet data classification}
To fully evaluate the effect of input mapping, comparison experiments are conduced on both Challenge 2012 and SantaFe data sets. Different from SantaFe data set described in Section IV-A, Challenge 2012 is a more complex and challenging data set. The goal of Challenge 2012 data is to predict mortality of 4000 Intensive Care Unit (ICU) patients using their 37 physiological signals (see Table \ref{ICU_feature}) which are measured at irregular time interval within total time length being 48 hours.  The irregularity of the measurement time interval and the complexity of the physiological signal make this prediction task highly challenging.

We first apply the following pre-processing procedure to the Challenge 2012:
\begin{itemize}
  \item Patient selection. The original data contains  records of 12000 ICU patients and the records are divided into two parts, training set A with 4000 patients' records and testing set B with 8000 patients' records. Because testing set B doesn't contain mortality state and thus we cannot calculate exactly its classification accuracy, we just conduct experiments on training set A. Some of the patients only have several measurements, for example due to an early in-hospital death. So we select the patients who have more than 40 records from set A and this results into a subset contains 3470 patients.
  \item Feature selection. The original data contain some features which have too few measurements during the observation period to enable effective temporal encoding, such as the TropT and TropI features are only measured several times during 48 hours. We first calculate the overall measurement amount of every feature and then select those feature with a average measurements more than 5. As a result the 12 features are used in the experiments, as listed in Table \ref{ICU_feature}.
\end{itemize}
\begin{table}[ht]
\caption{Physiological signals in Challenge 2012 data }\label{ICU_feature}
\centering  
\renewcommand\arraystretch{0.4} 
\scriptsize
\begin{tabular*}{8.5cm}{@{}ccc}\toprule \midrule
  Feature             &Meaning & Unit       \\ \midrule \midrule
DiasABP &Invasive diastolic arterial blood pressure & mmHg\\
GCS & Glasgow Coma Score (3-15)& -\\
HR &Heart rate & bpm\\
MAP & Invasive mean arterial blood pressure & mmHg\\
NIDiasABP & Non-invasive diastolic arterial blood pressure & mmHg\\
NIMAP & Non-invasive mean arterial blood pressure & mmHg\\
NISysABP & Non-invasive systolic arterial blood pressure & mmHg\\
RespRate &Respiration rate &bpm\\
SysABP &Invasive systolic arterial blood pressure &mmHg\\
Temp &Temperature & \grad C\\
Urine &Urine output & mL\\
Weight &-&kg      \\ \midrule \bottomrule
\end{tabular*}
\end{table}

\begin{figure}[ht]
  \centering
  \includegraphics[width=9cm]{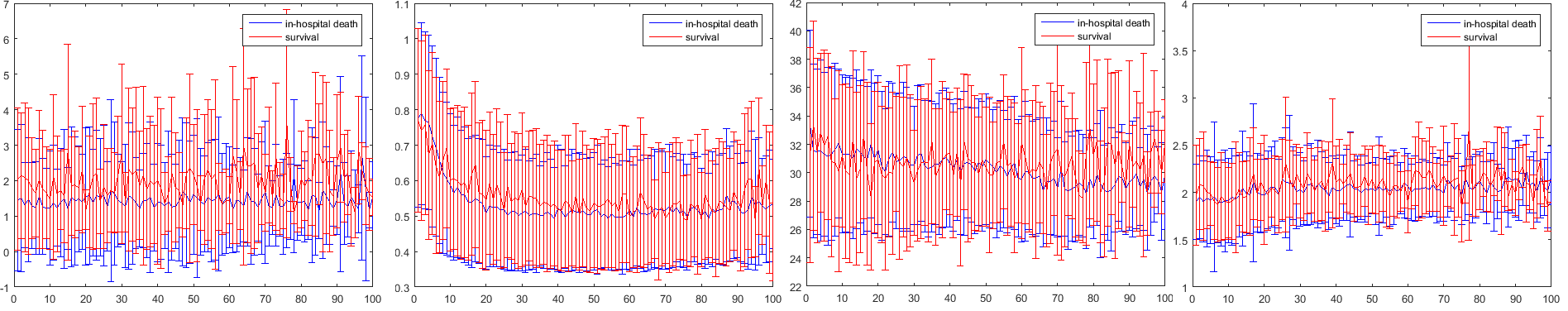}
  \caption{Mean and standard deviation of the first four selected features for in-hospital death and survival patients.From left to right: DiasABP, GCS, HR, MAP. Note how similar the signals of the two classes are.}\label{FeatureSignal}
\end{figure}
Figure \ref{FeatureSignal} displays the mean and standard deviation of the first four selected features for the two classes. From these figures we can see that the physiological signals corresponding to the two classes have very similar tends and the difference is almost  indistinguishable to human eyes \cite{bahadori2015functional}.

\begin{figure}[ht]
  \centering
  \includegraphics[width=4cm]{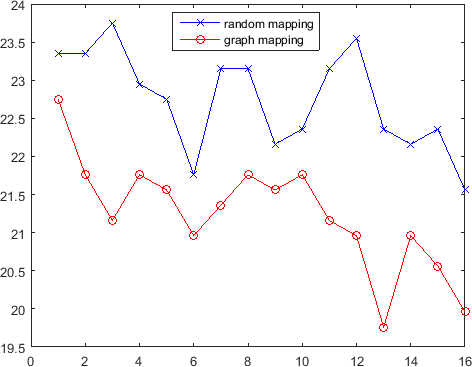}
\includegraphics[width=4cm]{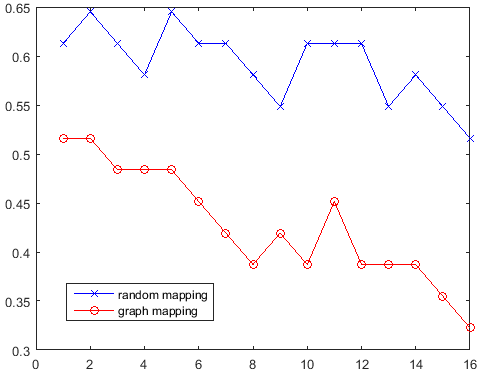}
  \caption{Error rate of random mapping and graph mapping on Challenge2012 data set (left) and SantaFe data set (right).}\label{GAOptimization}
\end{figure}

The comparison experiments are conducted in the following way. The GA optimization component is running with two modes: one is randomly mapping and the other mode is graph matching mapping. The optimized parameters are listed in Table \ref{OptParamTable}.
GA parameters are: generation number: 16, population size: 50, crossover function: scattered, crossover fraction: 0.2, selection function: roulette, elite count 5. The results Challenge 2012 and SantaFe are shown in Fig. \ref{GAOptimization}, in which horizontal axis represents GA generation and vertical axis represents error rate (fitness value of GA).
From these results we can see that the graph matching can achieve lower error rate than randomly mapping. As the generation number increases, we can see more clearly decreasing trend for graph based mapping, while the trend of random mapping curve is not so obvious and has greater fluctuation. It has to be mentioned that while the fitness value of GA algorithm decreases monotonically for optimizing deterministic system, it usually has fluctuation and does not always decrease for stochastic system. Because in a stochastic system even the system parameters are same, different initial states may yield different results. This is the reason why the error rate (fitness value of GA) in our NeuCube system dose not monotonically decrease.

\begin{figure}[ht]
  \centering
  \includegraphics[width=4cm]{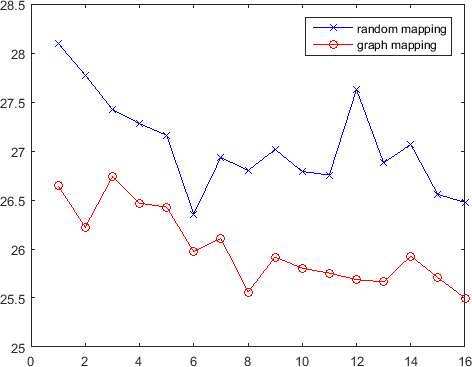}
  \includegraphics[width=4cm]{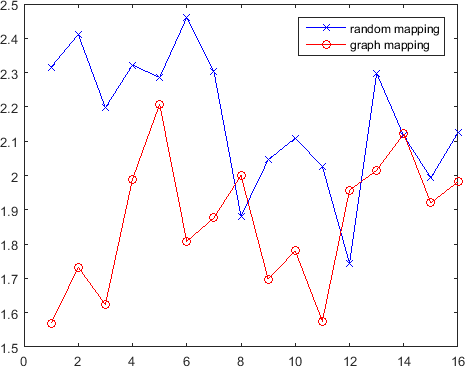}
  \caption{Mean (left) and standard deviation (right) of error rate of  random mapping and graph based mapping in GA parameter optimization for Challenge2012 data set.}\label{MeanStd}
\end{figure}

 To further compare the performance, Fig. \ref{MeanStd} displays the mean error rate and standard deviation of error rate of the 10 times running on Challenge 2012. From these results we can see the performance of graph based mapping method is more effective and stable, regarding to the lower mean error rate and the smaller standard deviation.

Since Challenge 2012 is a contest data set that highly challenges traditional classification algorithms, it might be interesting to compare the performance of NeuCbe with traditional methods. The results are shown in Table \ref{AccuracyTableICU}. Note the highest accuracy of baseline algorithms is 51.6\% from GAK, but the NeuCube can achieve 76.95\%, a significant improvement to state-of-art results. The optimal values of the parameters given by GA that attain this result are: Spike Threshold - 0.800,  Firing Threshold - 0.500,  STDP Rate - 0.073,   Refractory Time - 6,   Mod - 0.687,  Drift - 0.095.

\begin{table}[ht]
\caption{Comparative accuracy of spatiotemporal pattern classification (\%)}\label{AccuracyTableICU}
\centering  
\renewcommand\arraystretch{0.4} 
\scriptsize
\begin{tabular*}{8.8cm}{@{}cccccccc}\toprule \midrule
               &MLR     &SVM     &MLP        &$k$NN     &w$k$NN     &GAK             &NeuCube  \\ \midrule \midrule
Param.   & -            &$d$:2   &$n$:40       &$k$:6   &$k$:6           &$\sigma$:5 & -                                  \\ \cmidrule(r){2-8}
Acc.       &47.31     &46.81    &51.3      &49.3       &49.3           &51.6            &\textbf{76.95}                            \\ \midrule \bottomrule
\end{tabular*}

\end{table}


%
\section{Conclusions}
In this paper we proposed a new mapping method to map spatiotemporal input variables into a 3D spiking neural network architecture called NeuCube that enable NeuCube models any spatiotemporal data. The weighted undirect graph matching technique is adopted here so that similar input variables based on their temporal similarity are mapped into spatially closer neurons. The closer the neurons in the SNNcube are, the more temporal relationships they learn from data. This automatic optimal mapping algorithm greatly reduces the work load for a user to find out  the optimal mapping by hand in previous system and can yield better results regarding to spatiotemporal pattern recognition and early event prediction.  A comparison study of the proposed mapping and randomly mapping has been conducted on a popular contest physiological data set Challenge 2012 and the results demonstrated the superiority of the proposed method over randomly mapping method.

To have a better understanding of the SNNcube model and the data modelled in it, we also proposed an algorithm based on network activation spreading to automatically reveal neuronal clusters, which in previous system were also determined by hand according to the learnt  connection weights \cite{capecci2015analysis} and thus costed lots of time. The algorithm can divide the SNNcube into different neuronal clusters based on either the spike amount communication or the  connection weight between neurons. These neuronal clusters can help us to better understand the learning mechanism and the modelling ability of the SNNCube, as well as the data.

It should be mentioned that due to the scalability of the SNNcube, NeuCube is a \emph{deep} architecture and it has the potential to model any complex spatiotemporal data. Our future work includes:

\begin{itemize}
  \item Improvement of the mapping method for complex temporal and spatial data, including spatiotemporal data with moving spatial coordinates;
  \item Extension to including semi-supervised learning ability into the NeuCube model because semi-supervised learning has been demonstrated in various applications to be useful \cite{tu2013experimental, li2011semi, tu2015posterior} and is one of the basic ability of human neural system \cite{gibson2013human};
  \item Statistical analysis of  the firing pattern and spreading trajectory in SNNcube with mining algorithms \cite{tu2014novel} and how they reveal the significance of the SNNcube modelling ability;
  \item More experiments on other ecological-, environmental-, financial- and business  temporal data with large scale SNNcube to further explore the  deep modelling and deep prediction capability of the proposed NeuCube on complex spatiotemporal data.
\end{itemize}
\section*{Acknowledgment}
This work was funded by Education New Zealand and the Tripartite project between Auckland University of Technology, New Zealand, Shanghai Jiaotong University (East  China) and Xinjiang University (West China) China. The work was initiated in the Knowledge Engineering and Discovery Research Institute (KEDRI) during the visit of Enmei Tu in KEDRI and continued in the Shanghai JiaoTong University (SJTU) in a collaborative way. KEDRI and    SJTU  partially funded this work. This work is partly supported by NSFC China (No: 61572315),  and 973 Plan，China (No. 2015CB856004).

\ifCLASSOPTIONcaptionsoff
  \newpage
\fi



\bibliographystyle{IEEEtran}
\bibliography{refs}
%
%
%

%

\begin{IEEEbiography}[{\includegraphics[width=1in,height=1.25in,clip,keepaspectratio]{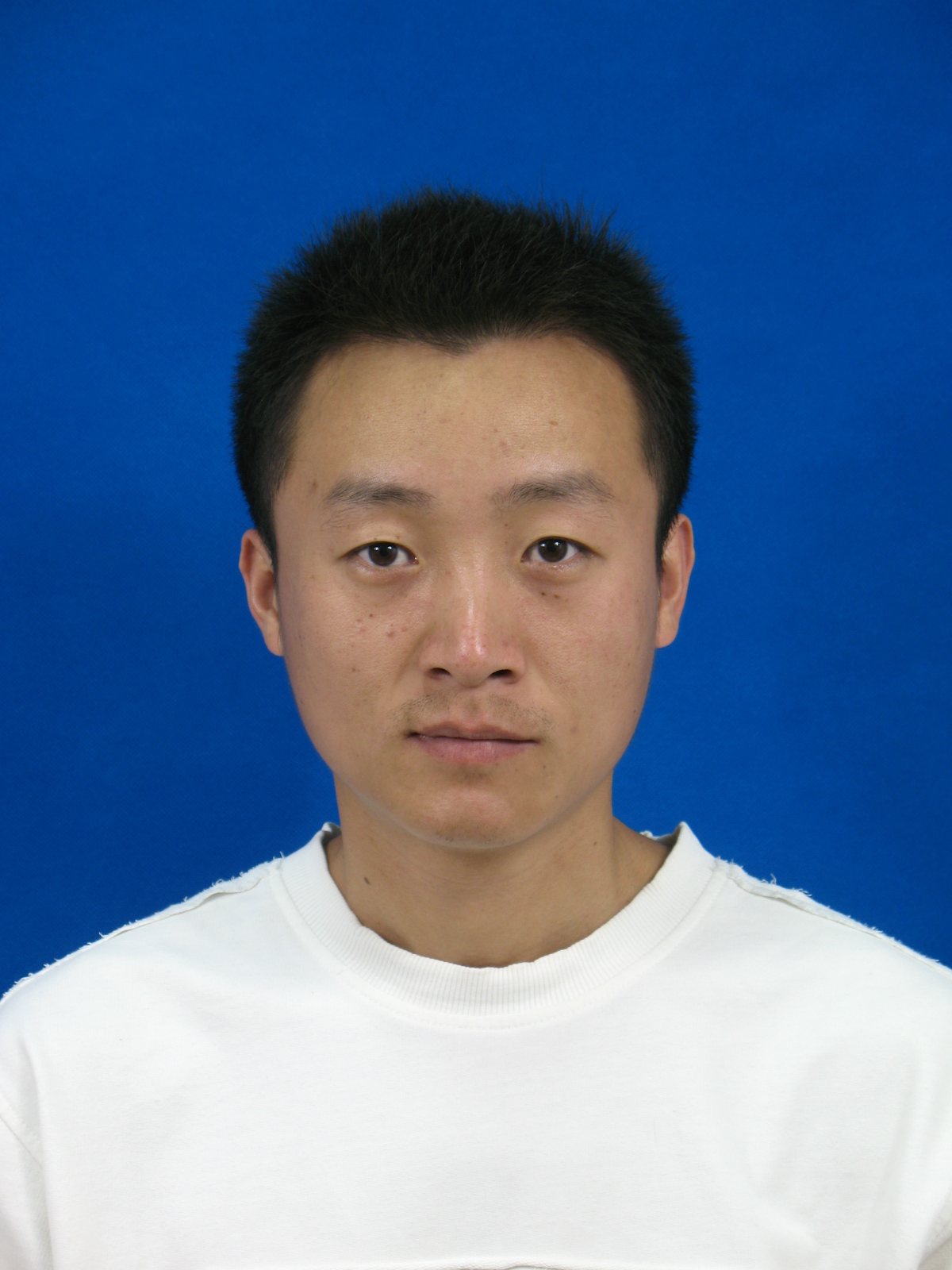}}]{Enmei Tu}
was born in Anhui, China. He received his B.Sc. degree and M.Sc. degree from University of Electronic Science and Technology of China (UESTC) in 2007 and 2010, respectively and PhD degree from the Institute of Image Processing and Pattern Recognition, Shanghai Jiao Tong University, China in 2014. He is now a research fellow in Roll-Royce@NTU Corporate Laboratory at Nanyang Technological University. His research interests are machine learning, computer vision and neural information processing.
\end{IEEEbiography}

\begin{IEEEbiography}[{\includegraphics[width=1in,height=1.25in,clip,keepaspectratio]{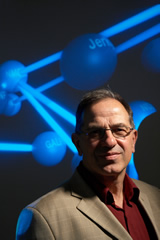}}]{Nikola Kasabov}
is a Fellow of the Royal Society of New Zealand, the New Zealand Computer Society and the Institute of Electrical and Electronic Engineers (IEEE). He is the founding Director and the Chief Scientist of the Knowledge Engineering and Discovery Research Centre (KEDRI) and Personal Chair of Knowledge Engineering in the School of Computing and Mathematical Sciences at AUT. His main interests are in the areas of: computational intelligence, neuro-computing, bioinformatics, neuroinformatics, speech and image processing, novel methods for data mining and knowledge discovery. He has published over 450 works in international journals and conferences, as well as books/chapters.
\end{IEEEbiography}

\begin{IEEEbiography}[{\includegraphics[width=1in,height=1.25in,clip,keepaspectratio]{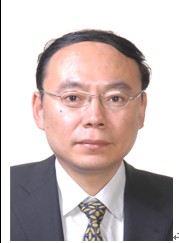}}]{Jie Yang}
received a bachelor’s degree and a master’s degree in Shanghai Jiao Tong University in 1985 and 1988, respectively. In 1994, he received Ph.D. in University of Hamburg, Germany. Now he is the Professor and Director of Institute of Image Processing and Pattern recognition in Shanghai Jiao Tong University. He is the principal investigator of more than 30 nation and ministry scientific research projects, including two national 973 research plan projects, three national 863 research plan projects, three national nature fundation projects, five international cooperative projects with France, Korea, Japan, New Zealand. He has published more than 5 hundreds of articles in national or international academic journals and conferences.
\end{IEEEbiography}

%
%
%




\end{document}